\documentclass[letterpaper]{article} 
\usepackage{aaai25}  
\usepackage{times}  
\usepackage{helvet}  
\usepackage{courier}  
\usepackage[hyphens]{url}  
\usepackage{graphicx} 
\usepackage{booktabs} 
\usepackage{amssymb}  
\usepackage{natbib}  
\usepackage{caption} 
\usepackage{multirow}
\usepackage{romanbar}  
\usepackage{algorithm}
\usepackage{algorithmic}
\usepackage{amsmath} 

\usepackage{newfloat}
\usepackage{listings}
\DeclareCaptionStyle{ruled}{labelfont=normalfont,labelsep=colon,strut=off} 
\floatstyle{ruled}
\newfloat{listing}{tb}{lst}{}
\floatname{listing}{Listing}
\title{Multi-path Exploration and Feedback Adjustment for Text-to-Image Person Retrieval}
\author{
    Bin Kang\textsuperscript{\rm 1},
    Bin Chen\textsuperscript{\rm 1\rm 2}\thanks{Corresponding Author.},
    Junjie Wang\textsuperscript{\rm 3},
    Yong Xu\textsuperscript{\rm 3}
}
\affiliations{
    \textsuperscript{\rm 1}School of Computer Science and Technology, University of Chinese Academy of Sciences, China\\
    
    \textsuperscript{\rm 2}International Research Institute of Artificial Intelligence, Harbin Institute of Technology (Shenzhen), China\\
         
    \textsuperscript{\rm 3}School of Computer Science and Technology, Harbin Institute of Technology (Shenzhen), China\\
   kangbin23@mails.ucas.ac.cn,       chenbin2020@hit.edu.cn
}

\begin{document}

\maketitle

\begin{abstract}
Text-based person retrieval aims to identify matching individuals using textual descriptions as queries. Existing methods largely depend on vision-language pre-trained models to facilitate effective cross-modal alignment. However, the inherent constraints of these models, including a proclivity for global alignment and limited adaptability, impede optimal retrieval performance. In response, we propose a Multi-path Exploration and Feedback Adjustment (MeFa) network, it utilizes a cyclical pathway of exploration, feedback, and adjustment to perform triple-refined associations both within and across modalities, thereby achieving more precise person-text association. Specifically, we devised an intra-modal reasoning pathway that generates challenging negative samples for cross-modal data to optimize intra-modal inference, thereby boosting model sensitivity to subtle variances. Subsequently, we introduced a cross-modal refinement pathway that leverages intermodal feedback to refine local information, thus enhancing its global semantic representation. Finally, the discriminative clue correction pathway incorporates fine-grained features of secondary similarity as discriminative clues, substantially enhancing the model's capacity to mitigate retrieval failures caused by disparities in fine-grained features. Experimental outcomes on three public TBPS benchmarks demonstrate that MeFa achieves superior cross-modal pedestrian retrieval without necessitating additional data or complex structures. The code will be made available upon acceptance.

\end{abstract}

%

\section{Introduction}

Text-based person retrieval task extends the person re-identification task \citep{Park_2020_AAAI, Ye_2021_TPAMI,Wang_2022_CVPR}, aiming to accurately retrieve and identify specific individuals from large-scale image databases based on textual descriptions as queries. By integrating vision-language pre-training(VLP) models \citep{Singh_2022_CVPR, Zhong_2022_CVPR, Li_2022_PMLR}, this task surpasses traditional image-to-image retrieval methods, allowing users to locate individuals using only text, thereby significantly enhancing the efficiency of personnel retrieval in complex environments. However, this strategy, which relies on VLP models \citep{Yu_2019_ICCV, Zhang_2018_ECCV}, also inherits the inherent limitations of these models, such as a preference for global information and poor self-regulation capabilities, making text-to-image person retrieval still challenging.

\begin{figure}[!t]
\centering
\includegraphics[width=\columnwidth]{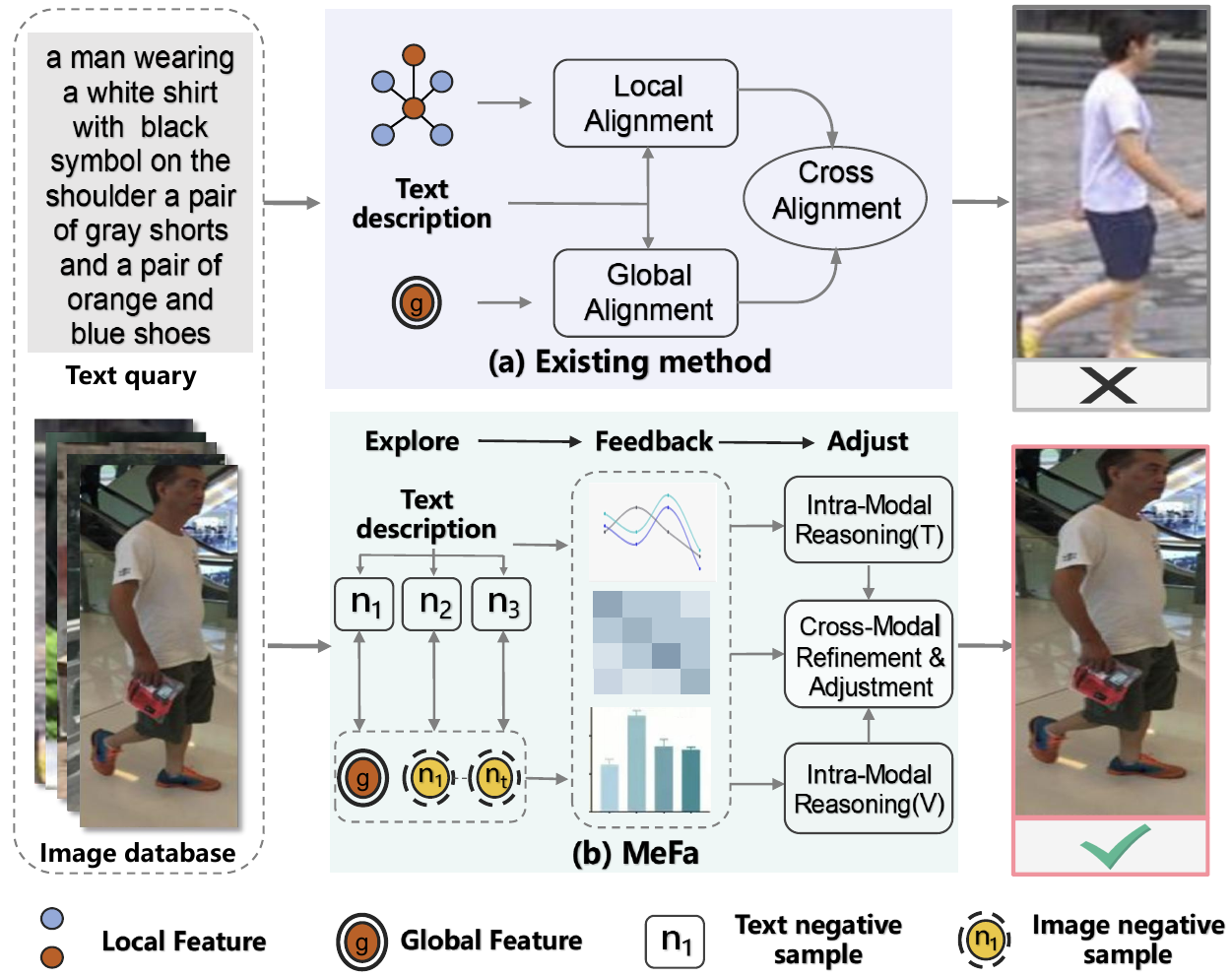} 
\caption{Illustration of MeFa Innovation: (a)Existing methods primarily rely on pre-trained models for feature extraction, using local, global, and cross-hard alignment techniques to explore correlations between textual descriptions and pedestrians. (b) MeFa enhances alignment accuracy and model robustness by constructing negative samples (IMR), refining features (CMR), and preserving discriminative clues (DCC).}
\label{fig1}
\end{figure}

In the realm of text-based person retrieval, it is imperative that models exhibit a nuanced comprehension of both coarse and fine-grained semantic features encapsulated in textual descriptors and their corresponding visual representations, thereby forging precise matching relationships across multimodal data. A principal challenge in this task is is the effective alignment of fine-grained features across visual and textual modalities. To address this, researchers have devised various sophisticated alignment mechanisms that capture global correspondences between whole person and textual descriptions \citep{Yu_2019_ICCV, Zhang_2018_ECCV} or achieve localized alignment between specific regions of person and detailed textual annotations \citep{Wang_2020_ECCV, Zhu_2021_MM}. However, such complex alignment strategies risk predisposing models to overfitting, significantly constraining their generalization capabilities in scenarios characterized by limited data or notable deviations in data distribution \citep{Yu_2019_ICCV, Zhang_2018_ECCV, Wang_2020_ECCV, Zhu_2021_MM}. Furthermore, some researchers have focused on extracting robust fine-grained feature sets, \citep{Shao_2022_MM}, while others have utilized auxiliary models to produce enriched textual descriptions that correspond closely to visual data \citep{Li_2022_PMLR, Bai_2023_MM}. While these methodologies bolster the model's ability to detect intricate details, they heavily depend on the quality of the generated or extracted fine-grained features and tend to neglect the broader semantic context \citep{Shao_2022_MM, Li_2022_PMLR, Bai_2023_MM}. This oversight can detrimentally affect the model’s efficacy when dealing with complex or ambiguous descriptions.

Recent research highlights the effectiveness of pretrained models such CLIP \citep{Radford_2021_pmlr} and ALBEF \citep{Li_2021_NEURIP}, which are trained on vast image-text datasets, in mastering cross-modal alignment and comprehending complex text and visual stimuli. Yan et al \citep{Yan_2023_TIP} advanced this field by proposing a multi-tiered alignment mechanism based on CLIP, significantly boosting retrieval performance. Further, the IRRA framework \citep{Jiang_2023_CVPR} capitalizes on these models to foster more effective visual-text alignment through implicit relational reasoning and cross-modal interactions. Despite these advantages, such models often falter in text-based person retrieval tasks that require discerning fine-grained distinctions, struggling to adapt to subtle image-text variations. This adaptation shortfall is twofold: 1) models such as CLIP and ALBEF become fixated on specific image-text pair processing during training, acking dynamic optimization capabilities from real-time feedback. 2) their application to fine-grained tasks frequently results in an inability to sift through and exclude irrelevant data associations, thus preventing adaptation to nuanced descriptive styles. Therefore, while VLP models demonstrate formidable cross-modal task handling, their limitations in alignment preferences and feedback regulation for specialized text-based retrieval tasks are critical and demand prompt resolution. These deficiencies significantly impair the models' operational performance and reliability.

In this paper, we propose a \textbf{M}ulti-path \textbf{e}xploration and \textbf{F}eedback \textbf{a}djustment (MeFa) network which employs a cyclical exploration, feedback, and adjustment pathway to progressively achieve finer granular information associations from intra-modal to inter-modal levels. Specifically, to boost the representational capabilities and convergence of the VLP model, we initially employ the efficient EvaCLIP\citep{sun_2023_evaclip} as the base model for extracting both image and text features. Building on this, we propose an intra-modal reasoning(IMR) pathway that utilizes feedback from negative samples highlighting subtle differences to adjust the model's internal representations, thereby improving its ability to discern fine distinctions. Subsequently, we designed a cross-modal refinement(CMR) pathway which intermodal interaction feedback refines global features into locally semantically-rich features, enhancing semantic consistency across scales. Lastly, we introduce a discriminative clue correction(DCC) pathway that utilizes fine-grained features of secondary similarity as discriminative cues to correct mismatches in person-text alignment caused by subtle differences. Contrasting with recent rigid alignment methods\citep{Cao_AAAI_2024}, MeFa depends on intrinsic feedback and adjustment across each pathway to ensure flexible and robust person-text matching. Our contributions are threefold:

\begin{itemize}

\item We propose MeFa network that employs an exploration, feedback and adjustment paradigm to mitigate global information biases and enhances self-regulatory capabilities in pre-aligned VLP models.
\item Leveraging this paradigm, we meticulously associate fine-grained information across three pathways: intra-modal reasoning, cross-modal refinement, and discriminative cue correction.
\item  Rigorous experiments on three benchmarks including CUHK-PEDES \citep{Li_2017_CVPR}, ICFG-PEDES \citep{Ding_2021}, and RSTPReid \citep{Zhu_2021_MM} show that MeFa outperformed prevailing advanced methods and achieved enhanced model convergence.

\end{itemize}

\section{Related Work}

\textbf{Text-based Person Retrieval}. Existing methods for Text-based Person Retrieval can be classified into two categories based on whether VLP models are utilized: the first either does not employ, or only uses language pre-training models in the text branch  \citep{Devlin_2018}. These methods rigorously examine multi-layered associations between images and text to explore diverse intra-modal and inter-modal alignment mechanisms, thereby achieving varied granularities of cross-modal alignment (global-global \citep{Ding_2021}, local-local \citep{Xiao_2021}, and global-local \citep{Wang_2020_ECCV}. These methods assume consistent semantic correspondence between modalities, the dynamic and variable relationships between text descriptions and images mean that static alignment mechanisms may fall short in diverse real-world scenarios. The second type relies on VLP models pre-trained on large-scale multimodal datasets \citep{Radford_2021_pmlr, Li_2021_NEURIP}, utilizing their substantial cross-modal semantic capabilities to achieve accurate text-image matching, even with vague or unconventional text descriptions. Liu et al. \citep{Liu_2021_ICCV} used a contrastive learning framework to transfer knowledge from large-scale generic image-text pairs to image search tasks, significantly surpassing previous methods. Subsequently, Wang et al.\citep{Wang_2020_ECCV} introduced a dual pre-trained modality framework to transfer CLIP knowledge, yet its dependence on predefined prompt templates restricted flexibility in managing unconventional or novel attributes. Yu et al. \citep{Yu_AAAI_2024} explored a text-free learning framework using CLIP for video-based person retrieval tasks. 
\begin{figure*}[!t]
\centering
\includegraphics[width=0.9\textwidth]{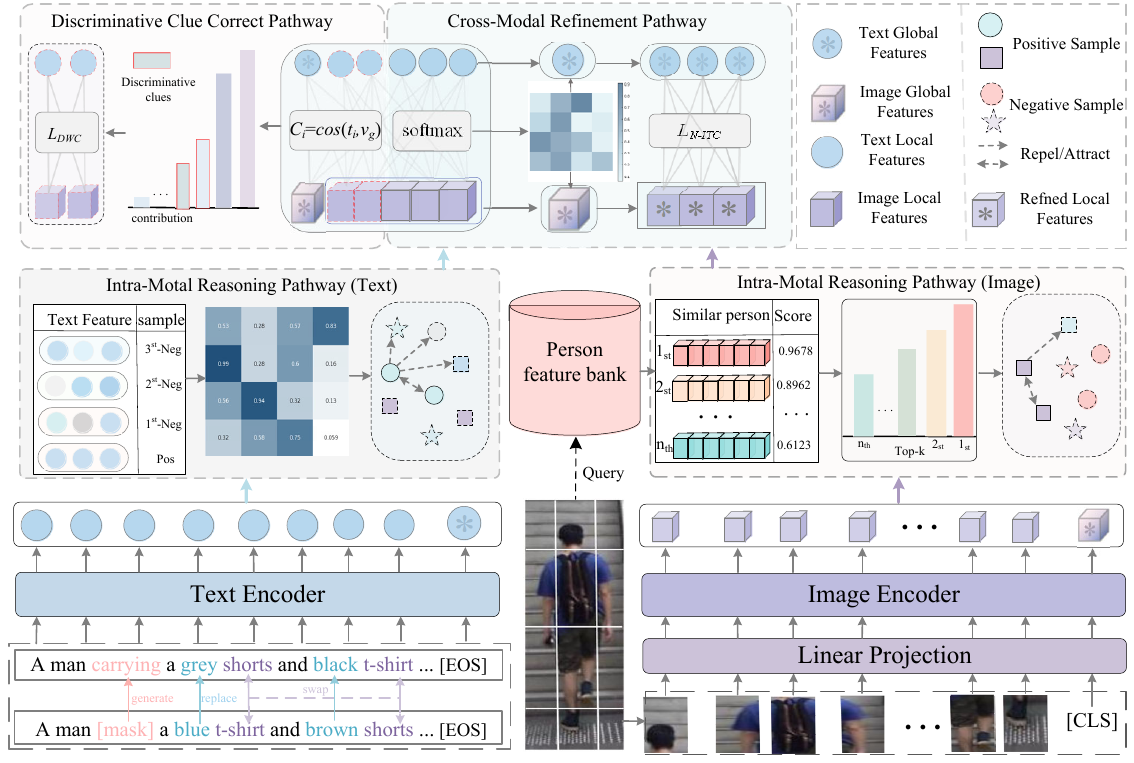} 
\caption{Overview of the proposed MeFa.It utilize the EvaCLIP encoder to extract textual and image features. The model's capability for fine-grained association is enhanced through three pathways: 1) Intra-modal inference pathway refines sensitivity to subtle variations using minimally different negative samples; 2) Cross-modal refinement pathway progressively transitions feature refinement from global to local; and 3) Discriminative clue correction pathway rectifies person-text mismatches caused by subtle discrepancies.}
\label{fig2}
\end{figure*}
These methods inherit inherent limitations of VLP models, including a preference for global information and poor self-regulation, which restricts model performance. In this work, we explore alignment through multiple pathways and adaptively adjust inter-modal associations based on feedback signals, proposing the MeFa framework to achieve more flexible and robust cross-modal fine-grained alignment.

\textbf{Vision-Language Pre-training} models have demonstrated significant potential across a variety of vision-language tasks. These tasks include image-text retrieval \citep{Diao_2023_TIP} and image captioning \citep{Yang_2023_CVPR}. By pre-training on vast collections of image-text pairs, these models acquire expressive multimodal feature representations that yield remarkable performance enhancements across several downstream tasks such as open-vocabulary detection \citep{Ma_NEURIPS2023, wang_2024}, zero-shot semantic segmentation \citep{Jiao_NEURIPS_2023}, and text-based pedestrian retrieval \citep{Liu_AAAI_2024, Jiao_NEURIPS_2023, Li_AAAI_2024}. A quintessential example of vision-language models is CLIP \citep{Radford_2021_pmlr}, which leverages 400 million internet-sourced image-text pairs with contrastive learning to align features in a unified space, extensively applied in real-world tasks \citep{Luddecke_2022_CVPR, Subramanian_2022_ACL, Sanghi_2022_CVPR}, notably zero-shot and few-shot challenges. Following CLIP, EvaCLIP \citep{sun_2023_evaclip} introduces enhanced initialization and optimization mechanisms, boosting zero-shot transfer capabilities and demonstrating efficiency in various zero-shot and few-shot tasks. Despite these advancements, these models often prioritize global semantic alignment, neglecting the association of fine-grained local information crucial for tasks demanding detailed contextual awareness such as text-based pedestrian retrieval. Furthermore, these models typically lack the self-regulation required to adapt to the complexities and variations of real-world scenarios.

\section{Method}

In this section, we present MeFa, a framework tailored for text-based person retrieval that focuses on multi-granular and multi-level alignment exploration, augmented by adaptive feedback regulation to enhance performance. The overall structure of MeFa is illustrated in Figure 2.

\subsection{Preliminaries}

In this study, we utilized a classic dual-encoder architecture to extract textual and image features separately. Text-based person retrieval models built on frameworks like CLIP primarily optimize distances between matched and unmatched image-text pairs through contrastive learning, so the model's performance is high sensitivity to the quantity and quality of samples within a batch. To boost the CLIP model's feature representation and hasten its convergence, we selected the superior performing EVA-CLIP as the initial model to improve cross-modal alignment potential. Specifically, for an input image \( I \in \mathbb{R}^{H \times W \times C} \), \( I \) is divided into \( N \) fixed-size, non-overlapping patches. These patches are linearly mapped and then fed into an image encoder, producing local embedding representations \(\{v_1, v_2, \dots, v_n\}\) and a global embedding representation \( v_{\text{cls}} \in \mathbb{R}^D \), where \( D \) is the dimension of the embeddings. Similarly, the input text \( T \), represented as \( T \in \mathbb{R}^{M \times D} \) with \( M \) as the number of tokens and \( D \) as the embedding dimension, is encoded into a series of token features \(\{T_1, T_2, \dots, T_M\}\) and a global feature \( T_{\text{cls}} \in \mathbb{R}^D \).

\subsection{Intra-modal reasoning path}

In the TBPS task, the primary objective is to accurately interpret and match the complex semantic content between images and texts, necessitating a model capable of discerning subtle semantic differences. To enhance this ability, we initially construct a series of challenging negative samples. For text, we develop three tiers of negative samples to test different aspects of semantic understanding: Tier-one uses the NLTK tool to swap key nouns, testing the model’s sensitivity to subject-object relations; Tier-two randomly substitutes verbs or adjectives to evaluate the model's understanding of actions and states; and tier-three obscures key words and utilizes a pre-trained RoBERTa model for completion, thereby heightening uncertainty and probing the model’s capacity to manage ambiguous contexts. For people images, given the unique challenges of the TBPS task, simple data augmentation techniques are insufficient to accurately mimic real person feature. Consequently, we utilize the target person’s image as a query input, searching within a vectorized training database for visually similar pedestrians, and selecting the top-k similar pedestrians as negative samples. 

Furthermore, we have designed an intra-modal compound loss mechanism to enhance the model's ability to distinguish challenging negative samples which comprises two components: the intra-modal separation loss is designed to ensure substantial differentiation between positive and negative samples, expressed as follows:
\begin{equation}
L_{\text{imr}} = \max(0, \alpha + D(f_a, f_p) - D(f_a, f_n))
\end{equation}
where \(f_a\), \(f_p\), and \(f_n\) represent the feature embeddings of the target image, the matching text description, and the non-matching text description, respectively. \(D\) denotes the cosine similarity, and \(\alpha\) is a margin ensuring adequate separation between positive and negative samples. The intra-modal contrastive loss is designed to enhance the model's sensitivity across a set of negative samples, defined as:
\begin{equation}
L_{\text{imc}} = \frac{1}{N} \sum_{i=1}^N \log \left(1 + \exp(\gamma (D(f_a^i, f_p^i)) - \min_{f_n \in \mathcal{N}_i} D(f_a^i, f_n))) \right)
\end{equation}
where \(\gamma\) is a scaling factor adjusting the sensitivity of the loss, \(N\) represents the batch size, and \(\mathcal{N}_i\) indicates a set of challenging negative samples corresponding to the \(i\)th sample.

\subsubsection{Cross-modal refinement path}

To further bridge the gap between task images and text descriptions, we have refined local features by exploring cross-modal interactions between global and local elements. Specifically,we first adjust the representation of local features through attention weights between cross modal local features, which can be formulated as follows:
\begin{equation}
\hat{v}_i = \frac{\exp(s(v_i, t_i))}{\sum_{i=1}^N \sum_{j=1}^M \exp(s(v_i, t_i))} \cdot v_i
\end{equation}
where $v_i$ represents the $i$-th local feature of the image, and $t_j$ represents the $j$-th local feature of the text. $s(v_i, t_j)$ is the cosine similarity between $v_i$ and $t_j$. Although the interaction-enriched local representations can encode more detailed clues, the global features condense contextual information and high-level semantics. Thus, we employ weighted global features to refine each modality's local features, enhancing the semantic capability of local features. More specifically:
\begin{equation}
\hat{v}_i' = (\hat{v}_i \oplus g) \odot \sigma(W_f \cdot (\hat{v}_i \oplus g) + b_f)
\end{equation}
\begin{equation}
\hat{t}_i' = (\hat{t}_i \oplus g) \odot \sigma(W_f \cdot (\hat{v}_i \oplus g) + b_f)
\end{equation}
where $\oplus$ denotes feature concatenation fusion, $\odot$ denotes element-wise multiplication, $\sigma$ is the tanh activation function, and $W_f$ and $b_f$ are learnable parameters. These operations not only strengthen the semantic associations between cross-modal local features but also enhance the semantic representation of local features. After enhancing the interaction of local features, we introduce the NITC to refine the fine-grained information alignment of cross-modal features, further ensuring a high level of semantic consistency between the image and text descriptions at both global and local levels. Specifically:
\begin{equation}
L_{\text{nitc}} = -\frac{1}{2N} \sum_{i=1}^N \left( \sum_{j=1}^M \left( p_{i,j}\log\hat{q}_{i,j} \right) + \sum_{j=1}^M \left( q_{i,j}\log\hat{p}_{i,j} \right) \right)
\end{equation}
Where $p_{i,j}$ and $\hat{p}_{i,j}$ respectively represent the true association probability between instance $i$ and text $j$ and the model's predicted association probability, which helps the model capture rich local details and align cross-modal semantics at a more granular level.

\subsection{Distinct cues correction path}

After interaction alignment via the CMR pathway, the model effectively retrieves most prominent cross-modal associative features. At this juncture, resolving target ambiguities hinges on subtler discriminative cues. To more robust region-text matching, we initially compute intermodal similarities \(s(v_i, t_i)\), selecting pairs with intermediate levels of similarity. These pairs are employed to formulate a corrective state \(R_k\), encapsulating subtle yet pivotal semantic links between regions and texts. Specifically, \(R_k\) is comprised of the top \(K\) words that are secondarily relevant to the overall sentence, focusing the model on subtle discriminative details. We then introduce a Discriminative Weighted Contrastive Loss (D-ITC):

\begin{equation}
L_{\text{ditc}} = -\sum_{i=1}^{N} \log \frac{\exp((q_j \cdot R_k) / \tau)}{\sum_{j=1}^{M} \exp((q_j \cdot v_i) / \tau)}
\end{equation}
where \( q_j \) represents textual features, \( v_i \) denotes regional features, and \( \tau \) is a parameter tuning the softmax function’s acuity. This methodology not only enhances prominent features but also refines the representation of more subtle attributes, thereby elevating region-text matching performance.

\section{Experiments}

\subsection{Benchmark Setup}
\textbf{Datasets}. We evaluated the efficacy of mefa across three benchmarks. \textbf{(1) CUHK-PEDES}, the first large-scale text-based pedestrian recognition benchmark, encompasses 13,003 individuals and 40,206 images, each equipped with two manually crafted descriptions, totaling 80,412 textual descriptions. Descriptions average at least 23 words in length, encompassing rich visual details, and the dataset is divided into training, validation, and test sets. \textbf{(2) ICFG-PEDES}, sourced from the MSMT17 dataset, focuses on identity descriptions and details compared to CUHK-PEDES, comprising 54,522 images and corresponding text descriptions covering 4,102 individuals, with each description averaging 37 words. \textbf{(3) RSTPReid}, also based on MSMT17, aims to address real-world challenges and includes 20,505 images and 41,010 text descriptions involving 4,101 individuals. Each individual is captured in 5 images taken by 15 cameras, with each image accompanied by two descriptions, each no fewer than 23 words.

\textbf{Evaluation Metrics}. In evaluating these datasets, Rank-K accuracy metrics, including Rank-1, Rank-5, and Rank-10, are primarily utilized. Rank-1 accuracy denotes the frequency with which the correct individual ranks first in retrieval results. Rank-5 and Rank-10 extend this criterion to measure appearances within the top five and ten results, respectively. Additionally, mean Average Precision (mAP) is another pivotal metric that quantifies the average precision of multiple queries, critically assessing performance in fine-grained retrieval tasks by reflecting the accuracy of ranked results.

\textbf{Implementation Details}. Training is conducted on 8 RTX-4090 GPUs with a total batch size of 80. To ensure a fair comparison, we employed the ViT-B/32 CLIP as our backbone network, with all input images uniformly resized to \(224 \times 224\) pixels and the maximum length of text token sequences set at 77 tokens. To accelerate convergence and extract richly representative features, we initialized parameters using the EvaCLIP pretrained model. Following EvaCLIP's guidelines, we adjusted weights using the LAMB optimizer. The model was trained over a total of 12 epochs, with the learning rate linearly increasing from \(1 \times 10^{-6}\) to \(1 \times 10^{-5}\).

\subsection{Benchmark Results}

In this section, we compare our method with several state-of-the-art Methods on CUHK-PEDES, ICFG-PEDES and RSTPReid benchmark datasets.

\textbf{Results on CUHK-PEDS dataset.}Table 1 presents our results on the CUHK-PEDES dataset. To ensure fair comparisons, we meticulously aligned our model initialization and baseline parameters with other methods. MeFa achieved Rank-1, Rank-5, and Rank-10 accuracies of 75.07\%, 91.01\%, and 93.28\%, respectively, surpassing all methods using the CLIP model as a foundation, including CFine[18], IRRA[19], VGSG[42], and TBPS-CLIP[21]. Furthermore, MeFa demonstrated significant superiority in the mAP metric, which emphasizes fine-grained retrieval performance, underscoring the efficacy of our network's tailored optimization for precise alignment.

Additionally, we evaluated MeFa's scalability by employing the more robust visual backbone, EVA02-L, as the base model. The results were impressive, with a transition from CLIP to EVA02-L yielding a performance increase of +2.5 ranks. This enhancement is attributed to EVA02-L's advanced visual representations, which enable more detailed semantic correspondences between images, essential for identifying nuanced pedestrian-text associations.

\begin{table}[h]
\centering
\setlength{\tabcolsep}{2.0pt} 
\begin{tabular}{c|c|cccc}
\hline
\textbf{} & \textbf{Methods} & \textbf{Rank-1} & \textbf{Rank-5} & \textbf{Rank-10} & \textbf{mAP} \\ \hline
  \multirow{5}{*}{\rotatebox{90}{w/ CLIP}} & CMMT(ICCV'21) & 57.10 & 78.14 & 85.23 & -- \\ 
 & NAFS(MM'22) & 61.50 & 81.19 & 87.51 & -- \\ 
 & IVT(ECCV'22) & 65.59 & 83.11 & 89.2 & -- \\ 
 & CTLG(TCSVT'23) & 69.47 & 87.13 & 92.13 & 60.56 \\ 
 & RaSa(IJCAL'23) & 76.51 & 90.29 & 94.25 & 69.38 \\ \hline
 
  \multirow{8}{*}{\rotatebox{90}{w/ CLIP}} & CFine(TIP'23) & 69.57 & 85.93 & 91.15 & -- \\ 
 & VGSG(TIP'23) & 71.38 & 86.75 & 91.86 & 67.91 \\ 
 & IRRA(CVPR'23) & 73.38 & 89.93 & 93.71 & 66.13 \\ 
 & TBPS-CLIP(AAAI'24) & 73.54 & 88.19 & 92.35 & 65.38 \\ 
 & IRLT(AAAI'24)  & 74.46 & 90.19 & 94.01 & -- \\ \cline{2-6}
 
 & CLIP(ViT-B/16) & 65.73 & 86.39 & 92.01 & 61.97 \\ 
 & EvaCLIP(baseline) & 66.51 & 86.94 & 92.77 & 63.09 \\
 & MeFa(CLIP) & 74.13 & 91.17 & 94.54 & 68.25 \\ 
 & MeFa(EvaCLIP) & \textbf{75.22} &\textbf{ 92.08} & \textbf{95.68} & \textbf{69.13} \\ \hline
\end{tabular}
\caption{Comparison with state-of-the-art methods on CUHK-PEDES}
\label{tab:methods}
\end{table}

\textbf{Results on Other Benchmarks.} Tables 2 and 3 respectively present our findings on the ICFG-PEDES and RSTPReid datasets, noted for their abundant task-oriented imagery and granular textual descriptions. Under uniform baseline and experimental conditions, our MeFa model markedly excelled beyond all CLIP-based competitors across every assessment metric, surpassing most methods that leverage advanced baselines and enhanced data.Specifically, concerning Rank-k metrics, our model registered increases of 2.3\% in Rank-1, 2.3\% in Rank-5, and 2.0\% in Rank-10 over the leading CLIP-based model, TBPS-CLIP. Notably, the MeFa model continued to demonstrate significant superiority on the mAP metric, focused on fine-grained matching accuracy. These findings highlight the MeFa's outstanding proficiency in managing intricate textual and visual data, confirming its effectiveness and reliability in detail-oriented tasks.

\begin{table}[h]
\centering
\setlength{\tabcolsep}{2.5pt} 
\begin{tabular}{c|c|cccc}
\hline
\textbf{} & \textbf{Methods} & \textbf{Rank-1} & \textbf{Rank-5} & \textbf{Rank-10} & \textbf{mAP} \\ \hline
  \multirow{5}{*}{\rotatebox{90}{w/o CLIP}} & SSAN(arxiv'21) & 54.23 &  72.63 & 79.53 & -- \\ 
 & IVT(ECCV'22) & 56.04 & 73.60 & 80.22 & -- \\ 
 & LGUR(MM'22) & 57.42 &  74.97 & 81.45 & -- \\ 
 & RaSa(IJCAL'23) & 65.28 & 80.40 & 85.12 & 41.29 \\
 & APTM(MM'23) & 68.51 &  82.99 & 87.56 & 41.22 \\  \hline
 
 \multirow{8}{*}{\rotatebox{90}{w/ CLIP}} & TP-TPS(arxiv'23) & 60.64 & 75.97 &  81.76 & 42.78 \\ 
 & CFine(TIP'23) & 60.83 & 76.55 &  82.42 & -- \\ 
 & IRRA(CVPR'23) & 63.46 & 80.25 &  85.82 & 38.06 \\ 
 & DCEL & 64.88 & 81.34 & 86.72 & -- \\ 
 & TBPS-CLIP(AAAI'24) & 65.05 & 80.34 & 85.47 & 39.83 \\ \cline{2-6}

 & CLIP(ViT-B/16) & 56.23 & 74.29 & 81.62 & 30.85 \\ 
 & EvaCLIP & 57.44 & 75.79 & 82.22 & 33.03 \\
 & MeFa(CLIP) & 66.93 & 82.17 & 86.26 & 40.74 \\ 
 & MeFa(EvaCLIP) & \textbf{67.42} & \textbf{83.08} & \textbf{86.63} & \textbf{42.78} \\ \hline
\end{tabular}
\caption{Comparison with state-of-the-art methods on ICFG-PEDES}
\label{tab:methods}
\end{table}

\begin{table}[h]
\centering
\setlength{\tabcolsep}{2.5pt}
\begin{tabular}{c|c|cccc}
\hline
\textbf{} & \textbf{Methods} & \textbf{Rank-1} & \textbf{Rank-5} & \textbf{Rank-10} & \textbf{mAP} \\ \hline
\multirow{6}{*}{\rotatebox{90}{w/o CLIP}} & DSSL(MM'21) & 32.43 &  55.08 & 63.19 & -- \\ 
 & SSAN(arxiv'21) & 43.50 &  67.80 & 77.15 & -- \\ 
 & IVT(ECCV'22) & 46.70 &  70.00 & 78.80 & -- \\ 
 & CAIBC(MM'22) & 47.35 & 69.55 & 79.00 & -- \\ 
 & RaSa(IJCAL'23) & 66.90 & 86.50 & 91.35 & 52.31 \\
 & APTM(MM'23) & 66.45 & 85.60 & 90.60 &  \\   \hline
 
\multirow{8}{*}{\rotatebox{90}{w/ CLIP}} & TP-TPS(arxiv'23) & 50.65 & 72.45 &  81.20 & 43.11 \\ 
 & CFine(TIP'23) & 50.55 & 72.50 &  81.60 & -- \\ 
 & IRRA(CVPR'23) & 60.20 & 81.30 &  88.20 & 47.17 \\ 
 & TBPS-CLIP(AAAI'24) & 61.95 & 83.55 & 88.75 & 48.26 \\ \cline{2-6}

 & CLIP(ViT-B/16) & 56.67 & 78.09 & 86.62 & 42.85 \\ 
 & EvaCLIP & 57.81 & 78.52 & 86.47 & 44.03 \\
 & MeFa(CLIP) & 63.93 & 85.63 & 90.73 & 50.37 \\ 
 & MeFa(EvaCLIP) & \textbf{64.12} & \textbf{85.99} & \textbf{90.82} & \textbf{52.89} \\ \hline
\end{tabular}
\caption{Comparison with state-of-the-art methods on RSTPReid}
\label{tab:methods}
\end{table}

\subsection{Further Extensions}

\textbf{Qualitative Results}: To better understand the advantages of MeFa, we compared its retrieval results with baseline methods and provided a detailed visualization in Figure 4. The MeFa model clearly excels in identifying and utilizing fine-grained information between text and pedestrian images. Notably, it demonstrates higher precision in resolving person recognition issues caused by subtle differences. For example, in Case 1, our model accurately focuses on the details of a person's shoes and identifies subtle mismatches between the shoes and the text description, thereby avoiding erroneous retrieval results. This sensitivity to detail is crucial in complex scenarios. In the more challenging Case 2, where the description involves specific actions such as "arms with another person," MeFa successfully determines whether the actions in the image match the description, highlighting its robust ability to capture complex action details.

\begin{figure}[t]
\centering
\includegraphics[width=\columnwidth]{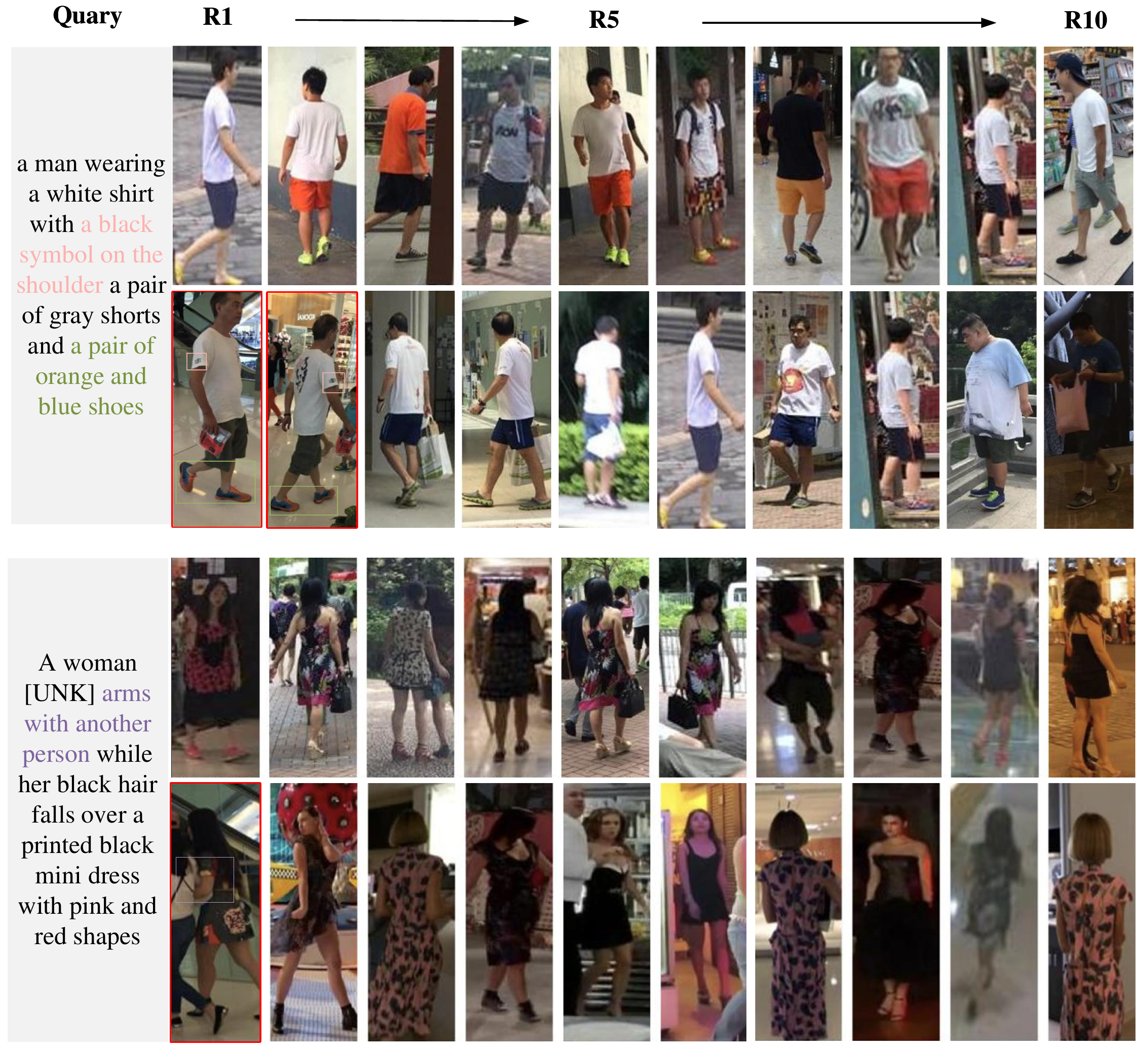} 
\caption{It displays Rank-10 qualitative retrieval results for MeFa versus baseline models, arranged in descending similarity from right to left. Correct matches are marked in red, with green and pink annotations highlighting precise details and actions successfully identified by MeFa in the image descriptions.}
\label{fig3}
\end{figure}

\textbf{Discriminative clue analysis}: Delving into discriminative cues is crucial for establishing fine-grained correspondences between modalities. Our analysis, illustrated in Figure 5, calculates the similarity between each word and person images, revealing significant findings. Common nouns such as "man," "shirt," and "shorts" demonstrate higher similarities with person images, mainly due to the direct associations established by the pretrained model between nouns and images. Conversely, articles and conjunctions like "the" and "a," despite their frequent textual occurrence, contribute minimally to similarity, thus offering limited discriminative power. However, adjectives and low-frequency nouns such as "white" and "symbol," which are vital for describing image details, only register secondary similarity values. This underrepresentation in overall similarity calculations can lead to the misidentification of individuals with similar primary features. To address this, we have opted to use these words as discriminative clues for secondary interactions, enhancing the distinction of difficult-to-distinguish samples prone to confusion.

\begin{figure}[t]
\centering
\includegraphics[width=\columnwidth]{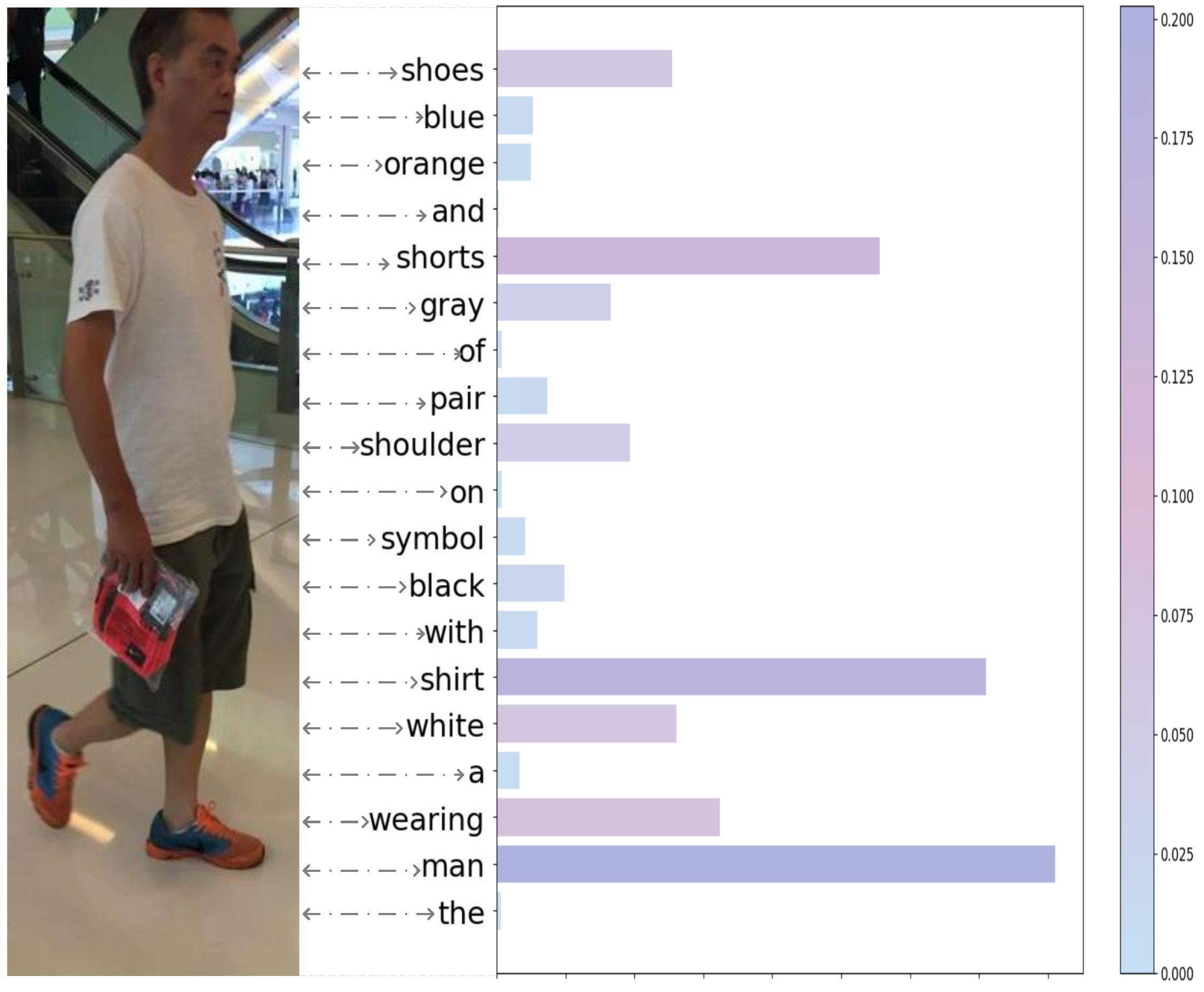} 
\caption{The similarity between part-of-speech segmented words and person images.}
\label{fig4}
\end{figure}

\subsection{Ablation}
\textbf{Component Efficacy Evaluation}: We conducted ablation studies on the CUHK-PEDES dataset to validate the individual contributions of each component within the MeFa framework, as detailed in Table 4. Initially using CLIP-ViT-16 as the baseline, we replaced the visual encoder with the enhanced EVA-CLIP, which improved Rank-1/5/10 accuracies by 7.67\%, 5.28\%, and 4.08\%, respectively, thereby establishing a robust baseline on CUHK-PEDES.Further exploration revealed that the intramodal inference pathway alone contributed to these improvements, with a notably greater enhancement at Rank-1 than at Rank-5 and Rank-10. This underscores the capability of constructing negative samples that highlight subtle differences, significantly boosting the model's sensitivity to such nuances. Additionally, the cross-modal refinement pathway and the discriminative information correction pathway considerably elevated the overall performance, particularly through the cross-modal refinement pathway which enhanced mAP performance by integrating global semantics with local information.Upon integrating all these components, the MeFa framework demonstrated the optimal retrieval performance, elevating overall accuracy from 65.56\% to 75.00\%. This substantial improvement not only confirms the effectiveness of each component but also illustrates their collective impact in enhancing the model's ability to process fine-grained information.

\textbf{Fine-Grained Information Effectiveness}: The accurate acquisition of fine-grained information is paramount for text-based person Retrieval. In order to evaluate MeFa's capability to mitigate the global information preference inherent in pretrained models, we executed a sequence of focused ablation experiments. Commencing from the preliminary retrieval results of the baseline model, we eliminated words delineating significant attributes (such as attire, pants, and backpacks) based on the similarity between images and tokenized words, maintaining the remainder of the text intact. Utilizing these revised texts alongside corresponding images, we methodically validated the contributions of each component. As evidenced in Table 5, the exclusion of critical feature descriptions precipitated a substantial decline in the baseline model’s Rank-1 index to 40.46\% and mAP to 37.4\%, accentuating the pretrained model’s inadequacies in handling intricate information. Nevertheless, with the stepwise incorporation of our model components, there was a 17\% enhancement in the Rank-1 index and a 14.62\% uplift in the mAP. relative to the baseline model dependent on coarse-grained original data, each element of MeFa markedly augmented the model’s capability to manage intricate details, effectively mitigating the global bias of pretrained models and conclusively substantiating MeFa’s effectiveness in augmenting the model’s proficiency in capturing subtle nuances.

\begin{figure}[t]
\centering
\includegraphics[width=\columnwidth]{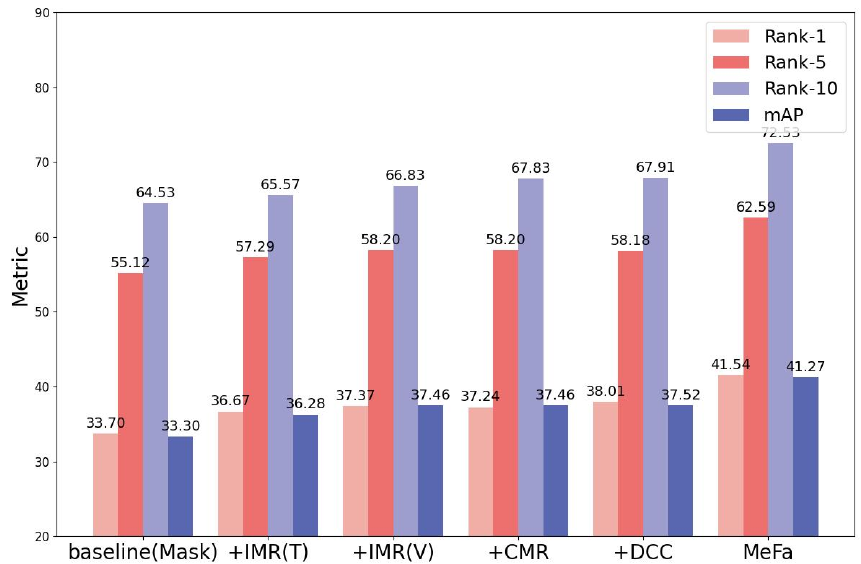} 
\caption{Ablation analysis of MeFa on fine-grained information. The baseline (Mask) comparison group demonstrates the effect of masking the top three high-frequency nouns in the test data, independently assessing the impact on fine-grained textual descriptions.}
\label{fig5}
\end{figure}

\newcommand{\myroman}[1]{\uppercase\expandafter{\romannumeral #1\relax}}

\begin{table}[htbp]
\centering
\caption{Ablation Study Results on the CUHK-PEDES Dataset}
\label{tab:ablation_study}
\renewcommand{\arraystretch}{1.2} 
\setlength{\tabcolsep}{1.5pt} 
\small 
\begin{tabular}{c|cccc|cccc}
\toprule
No. & IMR(T) & IMR(V) & CMR & DCC & Rank-1 & Rank-5 & Rank-10 & mAP \\ \midrule
0            &  --        &  --       & --          & --         & 66.51 & 86.94 & 92.77 & 63.09 \\  \hline
\myroman{1}  & \checkmark & --         & --         & --         & 67.73 & 87.39 & 93.01 & 61.97 \\
\myroman{2}  & --         & \checkmark & --         & --         & 68.80 & 88.06 & 93.39 & 62.99 \\
\myroman{3}  & --         & --         & \checkmark & --         & 69.87 & 88.73 & 93.77 & 64.02 \\
\myroman{4}  & --         & --         & --         & \checkmark & 70.94 & 89.40 & 94.15 & 65.04 \\  \hline
\myroman{5}  & \checkmark & \checkmark & --         & --         & 72.01 & 90.07 & 94.54 & 66.06 \\
\myroman{6}  & \checkmark & \checkmark & \checkmark & --         & 73.08 & 90.74 & 94.92 & 67.08 \\
\myroman{7}  & \checkmark & \checkmark & --         & \checkmark & 74.15 & 91.41 & 95.30 & 68.11 \\  \hline
\myroman{8}  & \checkmark & \checkmark & \checkmark & \checkmark & \textbf{75.22} &\textbf{ 92.08} & \textbf{95.68} & \textbf{69.13} \\
\hline
\end{tabular}
\end{table}

\subsection{Limitations and Conclusions}
In this paper, we focus on exploring the impact of global information preferences and self-regulation limitations in pre-trained models, introducing the MeFa network framework. By integrating modal internal reasoning, cross-modal refinement, and discriminative information correction, the MeFa network significantly enhances the model's accuracy in handling fine-grained information, achieving advanced results across various TBPR benchmarks. Moreover, our research identifies an interference effect when different alignment mechanisms are used together; the benefits brought by using alignment mechanisms A and B individually do not simply accumulate. This phenomenon suggests that the interactions between alignment mechanisms are complex and not yet fully understood. Consequently, we have deferred further exploration, awaiting more in-depth insights from future research.

\subsection{References}

\bibliography{aaai25}

\end{document}